\documentclass[conference]{IEEEtran}
\IEEEoverridecommandlockouts
\usepackage{cite}
\usepackage{amsmath,amssymb,amsfonts}
\usepackage{graphicx}
\usepackage{textcomp}
\usepackage{csquotes}
\usepackage{graphicx}
\usepackage{threeparttable}
\usepackage{dblfloatfix}
\usepackage{subcaption}
\usepackage{caption}
\usepackage{float}
\usepackage[utf8]{inputenc}
\usepackage{lipsum} %
\usepackage{mathrsfs} 
\usepackage[mathscr]{euscript}
\usepackage{bm}
\usepackage{hyperref}
\usepackage{xcolor}
\usepackage{enumitem} %
\usepackage[acronym]{glossaries}
\usepackage{csquotes}
\usepackage{tocloft} %
\usepackage{setspace}
\usepackage{amsmath}
\usepackage{amssymb}
\usepackage{algorithm}
\usepackage{algpseudocode}
\usepackage{graphicx}
\usepackage{verbatim}
\usepackage{stmaryrd}
\usepackage{tocloft} %
\usepackage{setspace}
\usepackage{makeidx} %
\usepackage{mathrsfs}
\usepackage{stmaryrd}
\usepackage{array}
\usepackage{booktabs}

\usepackage{todonotes}

\usepackage{tikz}
\usetikzlibrary{arrows.meta, positioning, shapes.geometric, shapes.symbols}

\def\BibTeX{{\rm B\kern-.05em{\sc i\kern-.025em b}\kern-.08em
    T\kern-.1667em\lower.7ex\hbox{E}\kern-.125emX}}

\newcommand{\reales}{\mathbb{R}}
\newcommand{\moyenne}{\mathbb{E}}

\newcommand{\boldtheta}{\boldsymbol{\theta}}

\newcommand{\matrices}[1]{\boldsymbol{#1}}
\newcommand{\tensors}[1]{\boldsymbol{\mathcal{#1}}}

\newcommand{\xbold}{\boldsymbol{x}}

\newcommand{\elbo}{\mathcal{L}_{\boldsymbol{\theta},\boldsymbol{\phi}}}

\newcommand{\boldphi}{\boldsymbol{\phi}}

\begin{document}
\title{Variational Low-rank Tensor Decomposition for Multisubject Spatiotemporal Data Analysis \\[-0.2em]}

\author{Laura M. Montaldo$^*$, Ricardo A. Borsoi$^*$, Sebastian Miron$^*$, Tülay Adali$^\dagger$ \\[0.6em]
$^*${\small CRAN, Université de Lorraine, CNRS, Vandoeuvre-lès-Nancy, France. email: firstname.lastname@univ-lorraine.fr.}\\[-0.2em]
$^\dagger${\small Department of CSEE, University of Maryland Baltimore County, Baltimore, MD 21250, USA. email: lastname@umbc.edu.} 
\thanks{This work was supported in part by the Agence Nationale de la Recherche (ANR) under projects ANR-23-CE94-0001, ANR23-CE23-0024, and by Université de Lorraine under LUE AAP INTERDISCIPLINAIRE 2022.}\\[-0.7em]
}

\maketitle

\begin{abstract}
Modeling shared and subject-specific structure in multisubject spatiotemporal data remains challenging, particularly in neuroimaging, where both spatial and temporal patterns exhibit rich variability across subjects. Existing matrix and tensor decompositions provide interpretable factorizations, but rely on fixed multilinear structures or coupling schemes that may limit their flexibility in capturing complex variability.
In this work, we introduce a spatiotemporal variational tensor decomposition (ST-VTD) framework that combines a tensor factorization generative model with structured priors to jointly represent spatial maps and temporal dynamics. Spatial factors are regularized to promote a low-rank structure inspired by the LL1 decomposition, while temporal factors are modeled using a learned Long short-term memory (LSTM)-based prior, enabling flexible and adaptive dynamics. Posterior inference is performed using an amortized variational formulation by unrolling iterations of an optimization algorithm, leading to an interpretable and parameter-efficient architecture. The proposed inference framework employs a warm-start strategy based on group independent component analysis, which we found to improve optimization performance. Experiments on a realistic synthetic functional MRI (fMRI) dataset demonstrate that the proposed approach significantly improves latent factor recovery compared with representative classical and probabilistic decomposition benchmarks.

\end{abstract}

\section{Introduction}
\label{sec:introduction}

Spatiotemporal data arise in a wide range of domains, including climate science, hyperspectral imaging, and functional neuroimaging~\cite{hannachi2007empirical,Blind_fMRI,liu2019reviewChangeDetectionHyperspectral}. 
In particular, in multisubject settings such as fMRI, one 
observes multiple datasets sharing common latent structures 
while exhibiting subject-specific variability. A central 
challenge is to decompose these data into spatial and temporal 
components that enable meaningful analysis and comparison across 
subjects.

Coupled matrix and tensor decompositions are widely used in spatiotemporal data analysis due to their interpretability and strong theoretical guarantees \cite{adali_reproducibility_2022,borsoi2024personalized,lahat_multimodal_2015,acar2015dataFusionCoupled,borsoi2021coupledBTD,chatzichristos2022coupledTensor}. However, this often comes at the price of imposing strong structural constraints on the spatial and temporal factors, which reduces their flexibility to represent complex relationships between datasets.

Classical approaches based on blind source separation and multilinear factorization, such as Group Independent Component Analysis (GICA)~\cite{calhoun2009review}, Canonical Polyadic Decomposition (CPD), and rank-$(L,L,1)$ decomposition (LL1)~\cite{kolda_tensor,sidiropoulos2017tensor}, as well as their coupled extensions~\cite{sorensen2015coupled,belyaeva2024spatiotemporalAdolescentCoupled}, extract global components shared across subjects. While effective for identifying common structure, these methods can be too restrictive to capture inter-subject variability.

More flexible tensor models have been proposed to relax these assumptions. 
In particular, PARAFAC2~\cite{kolda_tensor}, which can be efficiently estimated 
using alternating optimization schemes~\cite{parafac2_admm}, allows one factor to vary across 
tensor slices, while enforcing a structured form of variability. Similarly, coupled decompositions with shared and individual components~\cite{adali_reproducibility_2022, Borsoi_2023,borsoi2024personalized} introduce dataset-specific variability, but still require predefined coupling structures and often rely on post hoc alignment procedures. Other approaches, such as flexible coupling in CP models~\cite{farias2016cpdFlexibleCoupling}, enforce similarity between factors across datasets through distance-based regularizations.

From a probabilistic perspective, Bayesian CP factorization~\cite{bayesian_cp} improves robustness by enabling uncertainty quantification and automatic rank determination. However, it retains a CPD model with shared latent factors, limiting its ability to model subject-specific variability. %
Alternative statistical frameworks such as independent vector analysis (IVA)~\cite{adali_diversity_2014} introduce dependencies across datasets at the distributional level, but still rely on strong structural assumptions.

Alternative probabilistic formulations of factor analysis and tensor decomposition have also been investigated. Neural topographic factor analysis~\cite{neuraltopo} introduces a probabilistic generative framework for neuroimaging data with neural parameterizations that improve interpretability of spatial factors through structured latent factors. However, it provides limited modeling of temporal dependencies. %
A deep Markovian prior over temporal latent variables has also been incorporated to enable the modeling of nonlinear temporal dynamics while maintaining a structured spatial representation~\cite{farnoosh2021deep}.
Tensor decomposition models based on energy-based formulations~\cite{undirected} replace the standard generative model (which separates the prior and likelihood distributions) by a joint model over both observations and latent factors that can represent more complex data distributions. However, the interpretability of fully neural probabilistic parametrizations is limited.

Despite these advances, existing approaches either rely on restrictive assumptions on the latent factors or fail to jointly capture structured spatial variability and flexible temporal dynamics across multiple subjects. This motivates the development of models that combine probabilistic interpretability, expressive latent dynamics for spatiotemporal representations.

In this work, we propose a variational latent-variable framework for multisubject spatiotemporal data analysis, referred to as \emph{spatiotemporal variational tensor decomposition (ST-VTD)}. The model relies on a tensor factorization generative model to separate spatial and temporal latent factors, while promoting structured spatial representations through a low-rank prior. Temporal factors are modeled using a learned LSTM-based prior to capture complex temporal dependencies. This enables the decomposition to adapt across subjects while sharing information in a statistical formulation by means of common priors. 
An amortized variational inference framework is used for posterior inference by unrolling the iterations of a proximal gradient-based optimization algorithm \cite{LPALM}, leading to an interpretable and parameter-efficient architecture for the variational posterior that matches the prior assumptions. The proposed inference framework employs a warm-start strategy based on GICA, which we found to improve optimization performance.

Our contributions include: 1) a tensor factorization framework for spatiotemporal data analysis which allows the factors to adapt across subjects; 2) low-rank spatial priors and LSTM-based temporal priors that add complex structure to the model; 3) a parameter-efficient amortized variational framework for posterior inference leveraging unrolled optimization which matches modeling assumptions in the prior and posterior.

The remainder of the paper is organized as follows. Section~\ref{sec:model} introduces the proposed model and inference strategy. Section~\ref{sec:experiments} presents the data generation, baselines, and experimental results. Finally, Section~\ref{sec:cll} presents conclusions and directions for future work.

\paragraph*{Notation}
Scalars are denoted by lowercase letters ($x$), vectors by bold lowercase ($\xbold$), matrices by bold uppercase ($\matrices{X}$), and tensors by bold calligraphic symbols ($\tensors{X}$). The outer (tensor) product is denoted by $\circ$. 
We use the notation $\mathcal{N}_{\rm ew}(\matrices{A},\matrices{B})=\prod_{i,j} \mathcal{N}([\matrices{A}]_{i,j},[\matrices{B}]_{i,j})$ to represent Gaussian distributions of entrywise independent random variables. For a background on tensor decompositions, see~\cite{kolda_tensor,sidiropoulos2017tensor,comon2014tensorsReview}.

\section{Proposed Model and Inference Strategy}
\label{sec:model}

\subsection{Generative Model}
\label{sec:generative_model}

For each subject $n = 1,\dots,N$, we observe a spatiotemporal tensor $\tensors{X}^{(n)} \in \reales^{T \times V_x \times V_y}$, with $T$ time points and spatial resolution $V=V_x  V_y$,  where $V_x$ and $V_y$ denote the spatial dimensions along the horizontal and vertical axes, respectively. 
While we focus on order-3 tensors, higher-order data such as 4D fMRI volumes (containing \emph{width}, \emph{height} and \emph{depth} dimensions) \cite{Blind_fMRI} can be tackled by flattening two of the spatial modes to reshape it in the same  form as tensor $\tensors{X}^{(n)}$.

We model $\tensors{X}^{(n)}$ as a sum of outer products between spatial maps $\matrices{Z}_{k}^{(n)}\in\mathbb{R}^{V_x \times V_y}$ and temporal factors $\matrices{c}_{k}^{(n)}\in\mathbb{R}^{T}$:
\begin{align}
    \tensors{X}^{(n)} = 
    \sum_{k=1}^{K}
    \matrices{Z}_{k}^{(n)} \circ \matrices{c}_{k}^{(n)} + \tensors{E}^{(n)},
    \label{eq:bilinear}
\end{align}
with $\tensors{E}^{(n)} \sim \mathcal{N}_{\rm ew}(\boldsymbol{0}, \sigma^2\boldsymbol{1})$. By regrouping the temporal factors as $\matrices{C}^{(n)}=[\matrices{c}_{1}^{(n)},\ldots,\matrices{c}_{K}^{(n)}]$ and the spatial maps as a tensor $\big[\tensors{Z}^{(n)}\big]_{:,:,k}=\matrices{Z}_{k}^{(n)}$, for $k=1,\ldots,K$, we can write the data likelihood as
\begin{equation}
    p\big(\tensors{X} \mid \tensors{Z}, \matrices{C}\big) =
    \mathcal{N}_{\rm ew}\bigg(\sum_{k=1}^{K}
    \matrices{Z}_{k} \circ \matrices{c}_{k}, \, \sigma^2 \boldsymbol{1}\bigg),
    \label{eq:data_likelihood}
\end{equation}

where the subject index $(n)$ is omitted when denoting the random variable corresponding to a given subject observation, and $\boldsymbol{1}$ denotes a tensor of ones.  

To obtain an interpretable decomposition, it is crucial to introduce structural constraints as spatio-temporal priors, which we assume  factorize as 
$$p(\tensors{Z}, \matrices{C}) = \prod_{k=1}^K p(\matrices{Z}_{k}) p(\matrices{c}_{k}).$$

This is consistent with independence hypotheses widely used in source separation. For the spatial maps, based on the LL1 decomposition \cite{de2008decompositionsLL1Part2,domanov2020uniquenessLL1,prevost2022hyperspectralCoupledLL1,ding2023fast}, we consider an independent and identically distributed (i.i.d.) prior parametrized by a low-rank model as:
\begin{align}
    p(\matrices{Z}_{k}) =
    \mathcal{N}_{\rm ew}\big(\matrices{U}_{k} \matrices{V}_{k}^\top, \, \lambda_k^2\boldsymbol{1}\big), \quad k=1,\ldots,K \,,
\end{align}
where $\matrices{U}_k \in \mathbb{R}^{V_x \times L}$ and $\matrices{V}_k \in \mathbb{R}^{V_y \times L}$ are low-rank factors and $\lambda_k^2\in\mathbb{R}_+$ is a variance parameter per component shared across all datasets $n=1,\ldots,N$, and $L$ is a spatial rank parameter satisfying $L = \lfloor\min(V_x,V_y)/K\rfloor$, shared for all $k$ components. This low-rank prior reduces the number of spatial parameters while preserving local spatial structure, which is consistent with the smooth
activation patterns commonly observed in fMRI data. For the temporal factors, we consider a Gaussian parametrized by an LSTM network~\cite{lstm2014}:
\begin{align}
    p(\matrices{c}_{k}) =
    \mathcal{N}_{\rm ew}\big(\matrices{w}(\matrices{t}, \matrices{\theta}_{k}), \, \boldsymbol{\gamma}_k\big), \quad k=1,\ldots,K \,,
\end{align}
where $\matrices{t}=[1,2,\ldots,T]$ is a vector of time coordinates and $\matrices{w}(\matrices{t}, \matrices{\theta}_{k})$ represents a parametrization of the time courses as the output of an LSTM network, with $\matrices{\theta}_{k}$ being its parameters for component $k$, shared across all datasets and $\boldsymbol{\gamma}_k$ contains the variances, also shared across $n$. Unlike linear autoregressive priors, the proposed LSTM prior can represent nonlinear long-range temporal dependencies without imposing
a fixed temporal correlation model.

\subsection{Inference strategy}

Our goal is to infer the posterior  $p(\tensors{Z}, \matrices{C} \mid \tensors{X})$, which is intractable due to model nonlinearities and non-conjugate priors. We thus use variational inference with an approximate posterior distribution $q(\tensors{Z}, \matrices{C} \mid \tensors{X})$ which will be optimized to be close to the true posterior~\cite{bishop2006patternBook}. We design the variational distribution to have a similar structure as the generative model, in particular, by considering a conditional Gaussian mean field approximation structured as 
\begin{align}
    &\textstyle{q\big(\tensors{Z}, \matrices{C} \mid \tensors{X}\big)} = \textstyle{q\big(\tensors{Z} \mid \tensors{X}\big) q\big(\matrices{C} \mid \tensors{X}\big)}
    \nonumber \\
    & = \prod_{k=1}^K \mathcal{N}_{\rm ew}\big(\matrices{\mu}_{z,k}(\tensors{X}), \matrices{\vartheta}_{z,k}(\tensors{X})\big) \mathcal{N}_{\rm ew}\big(\matrices{\mu}_{c,k}(\tensors{X}), \matrices{\vartheta}_{c,k}(\tensors{X})\big),
    \nonumber 
\end{align}
where functions $\matrices{\mu}_{z,k}$ and $\matrices{\mu}_{c,k}$ (respectively,  $\matrices{\vartheta}_{z,k}$ and $\matrices{\vartheta}_{c,k}$) compute the means (respectively, entrywise variances) of the variational posterior based on the input data $\tensors{X}$.

To promote interpretability and a low number of parameters, rather than relying on a generic neural encoder, a central design choice is to exploit the generative model structure to compute the posterior means $\matrices{\mu}_{z,k}$, $\matrices{\mu}_{c,k}$. Thus, we use a network architecture obtained via algorithm unrolling~\cite{LPALM,kervazo2024deepUnrollingNMF,monga2021unrollingReviewSPM}. LPALM is particularly well suited to our setting because it naturally incorporates the low-rank projection associated with the spatial prior while preserving the interpretability of the optimization-based unrolled architecture.  Specifically, we unroll $I$ iterations of the Linearized Proximal Alternating Least-squares Minimization (LPALM)~\cite{LPALM} algorithm aimed at optimizing the loss function $\frac{1}{2} \|\matrices{X} - \matrices{C}\matrices{Z}^\top\|_F^2$, with the constraint that $\matrices{Z}$ has low-rank spatial structure.

Here $\matrices{X}\in\mathbb{R}^{T\times V}$, $\matrices{Z}\in\mathbb{R}^{V\times K}$ and $\matrices{C}\in\mathbb{R}^{T\times K}$ denote the observed tensor and the latent factors reordered as matrices. Starting from a trainable initialization $(\matrices{C}^{(0)},\matrices{Z}^{(0)})$, the temporal and spatial factors are updated at each iteration $\ell$ via gradient steps with step sizes $(\alpha_\ell, \gamma_\ell)$ as:
\begin{align*}
  \matrices{C}^{(\ell+1)}
  &= \matrices{C}^{(\ell)}
     - \alpha_\ell
       \bigl(\matrices{C}^{(\ell)}(\matrices{Z}^{(\ell)})^\top - \matrices{X}\bigr)
       \matrices{Z}^{(\ell)},
\end{align*}
and
\begin{align*}
  \matrices{Z}^{(\ell+1)}
  &= \mathcal{P}_{\rm LR}\!\left(
       \matrices{Z}^{(\ell)}
       - \gamma_\ell
         \bigl(\matrices{C}^{(\ell+1)}(\matrices{Z}^{(\ell)})^\top - \matrices{X}\bigr)^\top
         \matrices{C}^{(\ell+1)}
     \right),
\end{align*}
where $\mathcal{P}_{\rm LR}$ denotes a column-wise low-rank projection. Each column $\matrices{Z}_{:,k}$ is reshaped into $\matrices{Z}_k \in \mathbb{R}^{V_x \times V_y}$, projected onto rank-$L$ matrices via truncated SVD, and vectorized back, with $L=\lfloor\min(V_x,V_y)/K\rfloor$. This enforces a low-rank spatial structure consistent with the prior. The step sizes are set as 
$\alpha_\ell = \|\matrices{Z}^{(\ell)}\|_2^{-2}$ and 
$\gamma_\ell = (1.05\,\|\matrices{C}^{(\ell+1)}\|_2^2)^{-1}$, ensuring stable descent. The posterior log-variances are obtained from lightweight MLPs applied to the vectorized means:
$\log \boldsymbol{\vartheta}_{z} = \mathrm{MLP}_z(\mathrm{vec}(\boldsymbol{\mu}_z)) \in \mathbb{R}^K$ and
$\log \boldsymbol{\vartheta}_{c} = \mathrm{MLP}_c(\mathrm{vec}(\boldsymbol{\mu}_c)) \in \mathbb{R}^K$.
Both networks use three layers with widths $(VK,64,32,K)$ and $(TK,64,32,K)$, respectively, and ReLU activations.

\subsection{Optimization}

The training procedure consists of estimating the parameters of the prior and likelihood, which we regroup in $\theta$, and of the variational posterior, which we represent in $\phi$, by approximately maximizing the data likelihood (also called \emph{evidence})~\cite{murphy2018machineLearningBook}. This is performed by maximizing the evidence lower bound (ELBO), a surrogate lower bound on the intractable log-marginal likelihood, $\log p_{\boldtheta}(\tensors{X})\geq \elbo(\tensors{X})$ given by
\begin{equation}
\begin{aligned}
  \elbo(\tensors{X})
    :=\; & \moyenne_{q_{\boldphi}(\tensors{Z}|\tensors{X}) q_{\boldphi}(\matrices{C}|\tensors{X})}\!\left[
      \log p_{\boldtheta}(\tensors{X} \mid\tensors{Z},\matrices{C})
  \right] \\
  & - \mathrm{KL}\!\left(
      q_{\boldphi}(\matrices{C}\mid\tensors{X})
      \,\|\, p_{\boldtheta}(\matrices{C})
  \right) 
  \\& 
  - \mathrm{KL}\!\left(
      q_{\boldphi}(\tensors{Z}\mid\tensors{X})
      \,\|\, p_{\boldtheta}(\tensors{Z})
  \right),
\end{aligned}
\label{eq:elbo}
\end{equation}
where $\mathrm{KL}$ denotes the Kullback--Leibler divergence, and we make explicit the dependence of the probability density functions (PDFs) on their parameters. The first term in \eqref{eq:elbo} promotes data reconstruction, while the remaining terms enforce the posterior to agree with the structure of the prior. Importantly, maximizing the ELBO implies  minimizing  $\mathrm{KL}(q_{\phi}(\tensors{Z}, \matrices{C} \mid \tensors{X})\| p(\tensors{Z}, \matrices{C} \mid \tensors{X}))$ \cite{diederik2019introductionVAEs}.

The ELBO in~\eqref{eq:elbo} is empirically approximated over the dataset $\{\tensors{X}^{(n)}\}_{n=1}^N$, and maximized using stochastic gradient ascent over mini-batches of $B$ subjects, where the reparametrization trick~\cite{VAE_bayes} is used to sample from the variational distribution and estimate the gradients with respect to the parameters of both the generative model and the variational distribution. 
To improve the training efficiency, we also introduce annealing schedules for the temporal and spatial KL terms to mitigate posterior collapse~\cite{beta_VAE}, which is particularly important given the data might be imbalanced ($V \gg T$). 
The annealing coefficients $\beta_z(\tau)$ and $\beta_c(\tau)$ will weight the KL terms in the ELBO loss over the optimization iterations $\tau\in\mathbb{N}_+$ \cite{cyclical_annealing}. They are increased linearly from $0$ ($\tau=1$) to their maximum values $\beta_z^{\max}$ and $\beta_c^{\max}$ over $W_z$ and $W_c$ steps.

\section{Experiments}
\label{sec:experiments}

We evaluate the proposed method against benchmarks on a synthetic multi-subject fMRI dataset with significant inter-subject variability in the latent factors.

\subsection{Hybrid Synthetic Data}
\label{sec:synthetic_data}

\begin{figure*}[h!]
    \centering
    \includegraphics[width=0.65\textwidth]{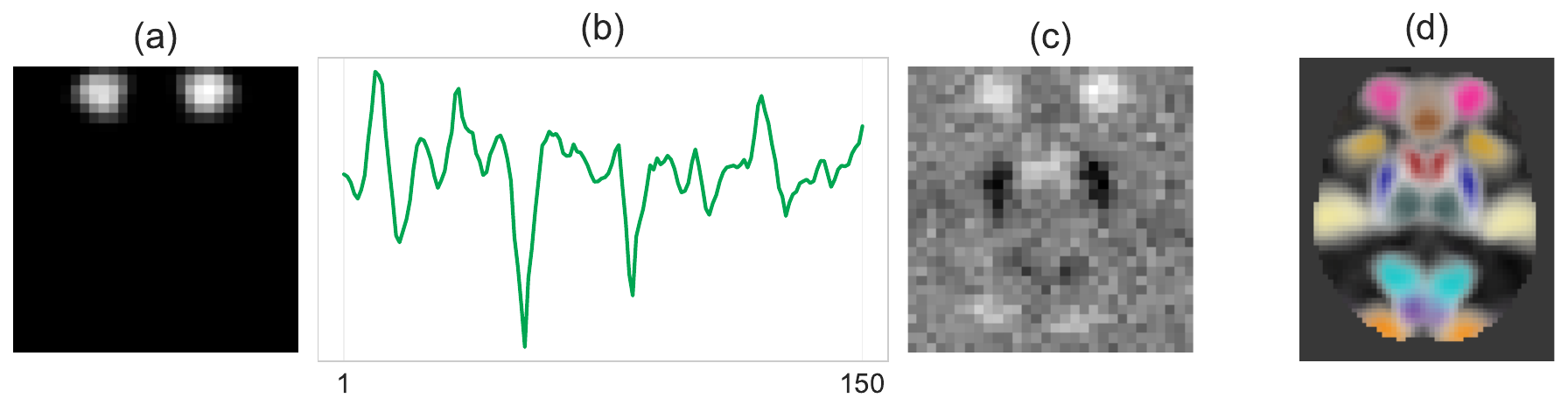}
    \caption{Ground-truth spatial map for component $k=7$ (a), its corresponding time course (b), the observed data slice at time $t=150$ for subject $n=1$ (c), and the 10 selected spatial maps overlaid on the data slice (d).}
    \label{fig:SM}
\end{figure*}

We construct a synthetic dataset that combines realistic anatomical structure with controlled inter-subject variability.  We select $K=10$ axial spatial maps from the Neuromark atlas~\cite{du2020neuromark}, shown in Fig.~\ref{fig:SM} (d), and use them as templates. Subject-specific spatial maps are generated by applying small perturbations (e.g., slight rotations), preserving anatomical structure while introducing inter-subject variability. The spatial intensities are further filtered using a percentile threshold, rescaled to lie in $[0,1]$ and reshaped.  In parallel, we generate $K=10$ smooth time courses (TC) per subject, with subject-dependent amplitude, phase shifts, and noise. 

For each subject, the $K$ ground-truth (GT) spatial and temporal components
are combined through the bilinear 
model in \eqref{eq:bilinear}, while introducing realistic perturbations including spatial deformations, amplitude variability, temporal phase shifts, and anatomical variability. White Gaussian noise with standard deviation $\sigma = 0.1$ is then added to obtain the observed data, resulting in spatiotemporal tensors
with $N=100$ subjects, $T=150$ time points, and $V_x = V_y = 30$ pixels.  In Fig.~\ref{fig:SM}(a)–(c), we show the ground-truth spatial map, its corresponding time course, and the observed data slice at time $t=150$ for subject $n=1$ and component $k=7$.

\subsection{Benchmark Models Setup}

We consider four benchmark decomposition methods representing complementary structural assumptions for multi-subject spatiotemporal analysis.

\paragraph{Group ICA (GICA)}
The centered subject-level data matrices $\big(\mathbf{X}^{(n)}\big)^\top \in \mathbb{R}^{V \times T}$ are concatenated along the temporal dimension to obtain $\mathbf{X}_g \in \mathbb{R}^{V \times (N\cdot T)}$, with $V=900$. A rank-$K$ PCA basis $\mathbf{U} \in \mathbb{R}^{V \times K}$ is computed via truncated SVD, FastICA \cite{fastICA} is applied to $\mathbf{U}^\top \mathbf{X}_g$, and the resulting spatial maps $\mathbf{Z}$ are back-projected to pixel space, column-normalized to fix scale ambiguity, and used to estimate subject-specific time courses by least squares as $\mathbf{C}^{(n)} = \mathbf{Z}^\dagger \mathbf{X}^{(n)}$, where $\mathbf{Z}^\dagger$ denotes the Moore--Penrose pseudoinverse of $\mathbf{Z}$.

\paragraph{Multilinear rank-$(L_r,L_r,1)$ decomposition}
For each subject, the centered data tensor $\tensors{X}^{(n)} \in \mathbb{R}^{T \times V_x \times V_y}$ was first normalized by its Frobenius norm and then decomposed using the Tensorlab LL1 model~\cite{tensorlab30} with rank $R=K=10$ and block size $L_k=3$ for all components, satisfying the constraint $R\cdot L_k \le \min(V_x,V_y)$. The decomposition was estimated with the nonlinear least-squares solver, with random initialization. 
The resulting spatial and temporal factors were then rescaled to the original data norm and stored as subject-specific spatial maps $\mathbf{Z}^{(n)} \in \mathbb{R}^{V \times K}$ and temporal courses $\mathbf{C}^{(n)} \in \mathbb{R}^{T \times K}$.

\paragraph{Bayesian CP factorization (BCPF)}
For each subject, the centered input tensor was decomposed with the BCPF model~\cite{bayesian_cp} for fully observed tensors, using a fixed rank $R=K$. The model was initialized with the maximum-likelihood SVD-based scheme and run without automatic rank pruning. Noise precision was updated during inference. 
The resulting factors were used to recover subject-specific spatial maps and time courses.

\paragraph{PARAFAC2}
Temporal factors are allowed to vary across subjects, while the spatial maps are shared. The subject-level data matrix $\mathbf{X}^{(n)} \in \mathbb{R}^{T \times V}$ is modeled as $\mathbf{X}^{(n)} \approx \mathbf{C}^{(n)} \, \mathrm{diag}(\mathbf{a}^{(n)}) \, \mathbf{Z}^\top$, where $\mathbf{C}^{(n)} \in \mathbb{R}^{T \times K}$ contains the subject-specific temporal factors, $\mathbf{Z} \in \mathbb{R}^{V \times K}$ is the shared spatial factor, and $\mathbf{a}^{(n)} \in \mathbb{R}^{K}$ is a subject-specific scaling vector. The model was fitted using the algorithm in~\cite{parafac2_admm}, with rank $K=10$, default initializations, and up to 500 iterations.

\subsection{ST-VTD Setup and Ablations}

We implemented ST-VTD using $I=50$ LPALM iterations. Both the spatial
factors $\matrices{Z}^{(0)}$ and the temporal factors $\matrices{C}^{(0)}$
were initialized from GICA results. To allow the network to adapt the
starting point during training, a small subject-shared learnable
offset (initialized from $\mathcal{N}(0,0.01^2)$ and optimized jointly
with all other parameters) is added to each initialization, so that the
effective starting point is $\matrices{Z}^{(0)} = \matrices{Z}^{(0)}_{\rm
GICA} + \Delta_z$ and $\matrices{C}^{(0)} = \matrices{C}^{(0)}_{\rm GICA}
+ \Delta_c$, where $\Delta_z \in \mathbb{R}^{V \times K}$ and $\Delta_c \in
\mathbb{R}^{T \times K}$ are shared across subjects. All remaining
parameters were initialized randomly.

The model was optimized using Adam with a cosine annealing schedule
(minimum learning rate $10^{-6}$) and gradient clipping at norm $3.0$.
To reflect the different roles of each parameter group, we used
group-specific learning rates: $3\times10^{-5}$ for the encoder,
$3\times10^{-4}$ for the temporal prior, and $3\times10^{-6}$ for the
spatial prior. To ensure stable training, the log-variance outputs of the
multilayer perceptron (MLP) heads are clamped to $[-6, 2]$ and their final-layer biases are
initialized to $-6$, corresponding to a small initial variance.
Training ran for 150 epochs with mini-batches of 10 subjects,
$\beta_z^{\max}=\beta_c^{\max}=5$, and a linear KL warm-up over the first
$W_z=W_c=50$ optimization steps to avoid posterior collapse early in training. \\
The source code used to reproduce the experiments is publicly available on
\href{https://github.com/lmontaldo/stvtd}{https://github.com/lmontaldo/stvtd}.

To assess the impact of different modeling choices, we conducted an 
ablation study comprising 8 configurations obtained by combining:
1) two temporal priors: the proposed LSTM-based prior with learned time-varying mean and variance, versus a fixed standard normal prior $\mathcal{N}(\mathbf{0},\mathbf{I})$;
2) two spatial priors: the proposed low-rank prior with mean $\mathbf{U}_k \mathbf{V}_k^\top$ and a single shared variance per component, versus an unconstrained Gaussian prior with learned mean and per-voxel log-variance;
3) two spatial projection schemes $\mathcal{P}_{\mathrm{LR}}$: hard SVD truncation to the top-$L_k$ singular values, versus no projection.

Among these configurations, the combination of the LSTM temporal prior and the low-rank spatial prior with projection achieves the best component recovery performance. The corresponding best-performing configuration, whose results are reported in Table~\ref{tab:results}, is used in all subsequent experiments. The final model comprises approximately 19{,}300 trainable parameters and is trained on 90 subjects for 150 epochs in about 6 minutes on CPU. For comparison, a spatially coupled LL1 decomposition ($L=3$) has 151,800 parameters.

To explore the role of initialization of the trainable $(\matrices{C}^{(0)},\matrices{Z}^{(0)})$, we retrained the best configuration 
using random Gaussian initialization instead of GICA estimates. Performance 
degraded substantially: $\mathrm{re}_X$ increased by $12.5\%$, component 
recovery errors ($\mathrm{re}_{z}$, $\mathrm{re}_{c}$) increased on average 
by $\approx 335\%$, and spatial and temporal correlations dropped by $\approx  30\%$, 
illustrating the importance of initialization due to the non-convexity of the 
optimization problem.

\subsection{ST-VTD vs. Benchmark Models: Performance Comparison}

\begin{figure*}[!t]
    \centering
    \includegraphics[width=0.65\textwidth]{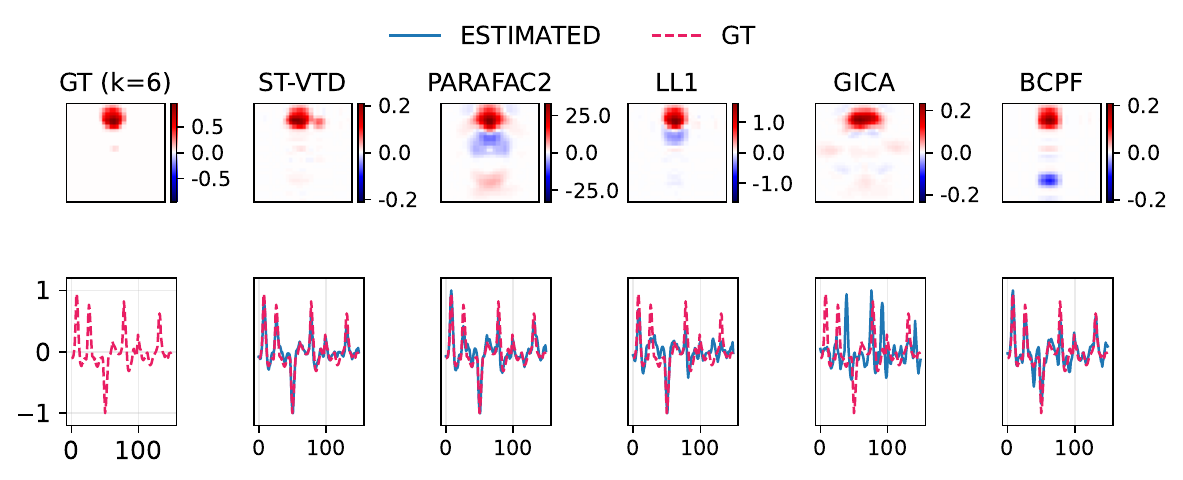} \\
    \vspace{-0.2cm}
    \caption{Ground truth (GT) and estimated component ($k=6$) for the same subject across all models.}
    \label{fig:stvoe_component9}
\end{figure*}

Let $\mathbf{X}^{(n)} \in \mathbb{R}^{T \times V}$ denote the observed data 
for subject $n$ and $\hat{\mathbf{X}}^{(n)}$ its reconstruction. Let 
$\mathbf{Z}_k^{(n)}, \mathbf{c}_k^{(n)}$ and 
$\hat{\mathbf{Z}}_k^{(n)}, \hat{\mathbf{c}}_k^{(n)}$ denote the true and 
estimated spatial maps and time courses, respectively. 
The estimated and true components are matched using the Hungarian algorithm and scaling to compensate for permutation and scaling ambiguities.
We evaluate performance using three metrics: reconstruction error ($\mathrm{re}_X$):
\begin{equation}
{\textstyle \mathrm{re}_X =
\frac{1}{N}\sum_{n=1}^N
\frac{\|\hat{\mathbf{X}}^{(n)} - \mathbf{X}^{(n)}\|_F}
     {\|\mathbf{X}^{(n)}\|_F}\,},
\end{equation}

mean absolute Pearson correlation over subjects and components ($\mathrm{corr}_{z}$, $\mathrm{corr}_{c}$):
\begin{equation}
    {\textstyle\mathrm{corr}_{z} = \frac{1}{NK}\sum_{n,k}
    \left|
    \frac{(\hat{\mathbf{z}}_k^{(n)} - \bar{\hat{\mathbf{z}}}_k^{(n)})^\top
          (\mathbf{z}_k^{(n)} - \bar{\mathbf{z}}_k^{(n)})}
         {\|\hat{\mathbf{z}}_k^{(n)} - \bar{\hat{\mathbf{z}}}_k^{(n)}\|_2\,
          \|\mathbf{z}_k^{(n)} - \bar{\mathbf{z}}_k^{(n)}\|_2}
    \right| \,},
\end{equation}

with $\mathbf{z}_k^{(n)} \in \mathbb{R}^{V}$ and 
$\mathbf{c}_k^{(n)} \in \mathbb{R}^{T}$, and analogously for $\mathrm{corr}_{c}$, and scale-invariant relative error ($\mathrm{re}_{z}$, $\mathrm{re}_{c}$):

$$
{\textstyle
\mathrm{re}_{z}=
\frac{1}{N}\sum_{n=1}^N
\frac{\left\|
\operatorname{diag}(\boldsymbol{\alpha}^{(n)})
\hat{\mathbf{Z}}^{(n)}
-\mathbf{Z}^{(n)}
\right\|_F}
{\left\|\mathbf{Z}^{(n)}\right\|_F}},
$$

where each element of $\boldsymbol{\alpha}^{(n)} \in \mathbb{R}^K$ is computed as 

$$
{\textstyle
\alpha_k^{(n)}=
\frac{
\left\langle
\hat{\mathbf{z}}_k^{(n)},
\mathbf{z}_k^{(n)}
\right\rangle}
{\left\|
\hat{\mathbf{z}}_k^{(n)}
\right\|_2^2}},\; \text{ for } \: k=1,\ldots,K
$$

and analogously for $\mathrm{re}_{c}$.

Table~\ref{tab:results} reports the performance metrics of all evaluated models. 
At evaluation time, reconstructions are computed from posterior means, corresponding to MMSE estimators.
ST-VTD achieves the best overall performance. It attains the lowest 
reconstruction error, with a negligible difference compared to LL1 
($0.2524$ vs.\ $0.2527$), and outperforms all baselines in latent recovery. 
In particular, it yields the highest spatial and temporal correlations 
($\mathrm{corr}_{z}=0.9829$, $\mathrm{corr}_{c}=0.9845$) and the lowest 
factor errors ($\mathrm{re}_{z}=0.1578$, $\mathrm{re}_{c}=0.1605$). 
Compared with GICA and PARAFAC2, it reduces reconstruction error by more 
than $50\%$ and improves factor agreement.

\begin{table}[h!]
\centering
\footnotesize
\setlength{\tabcolsep}{5pt}
\renewcommand{\arraystretch}{1.15}

\caption{Summary of test-set performance across methods}
\label{tab:results}

\resizebox{\columnwidth}{!}{
\begin{tabular}{lccccc}
\toprule
 & ST-VTD & LL1 & BCPF & GICA & PARAFAC2 \\
\midrule
$\mathrm{re}_X$
  & $\mathbf{25.2 \pm 0.7}$
  & $25.3 \pm 0.7$
  & $35.8 \pm 1.3$
  & $54.1 \pm 2.7$
  & $56.3 \pm 2.7$ \\
$\mathrm{corr}_{z}$
  & $\mathbf{98.3 \pm 2.6}$
  & $87.4 \pm 24.9$
  & $93.0 \pm 7.1$
  & $81.3 \pm 15.9$
  & $73.1 \pm 17.5$ \\
$\mathrm{corr}_{c}$
  & $\mathbf{98.5 \pm 3.5}$
  & $83.8 \pm 30.7$
  & $95.9 \pm 11.6$
  & $87.4 \pm 25.6$
  & $74.9 \pm 31.8$ \\
$\mathrm{re}_{z}$
  & $\mathbf{15.8 \pm 4.3}$
  & $40.0 \pm 11.0$
  & $35.2 \pm 3.9$
  & $52.5 \pm 2.8$
  & $63.0 \pm 2.5$ \\
$\mathrm{re}_{c}$
  & $\mathbf{16.0 \pm 6.1}$
  & $43.2 \pm 12.9$
  & $25.0 \pm 6.5$
  & $38.8 \pm 14.4$
  & $57.7 \pm 7.1$ \\
\bottomrule
\end{tabular}
}

\vspace{0.15cm}
\footnotesize
All values are multiplied by 100. For relative errors, lower is better; for correlations, higher is better. The best result in each row is shown in \textbf{bold}.
\end{table}

Fig.~\ref{fig:stvoe_component9} shows the ground truth (GT) and the estimated 
component $k=6$ corresponding to the subject for which ST-VTD achieved the median reconstruction error ($n=1$ across models). 
The first column displays the true SM and TC, and the remaining columns the estimates 
for each method. PARAFAC2, BCPF, and LL1 exhibit artifacts with large negative 
regions that distort the spatial structure, while GICA produces smoother but 
more diffuse estimates with increased leakage. In contrast, ST-VTD yields a 
spatial map that better matches the ground truth, with reduced leakage and 
improved fidelity. Although a small adjacent artifact remains, its 
amplitude is low, leading to a more accurate overall reconstruction.

\section{Conclusion and Discussion}
\label{sec:cll}

ST-VTD achieved the best overall trade-off among the compared methods, matching the strongest reconstruction performance while clearly improving latent-factor recovery. This suggests that the structured variational design is effective not only for reconstruction, but also for capturing more faithful spatiotemporal representations. In practice, the LPALM-based encoder is still sensitive to initialization, and the GICA warm start proved beneficial for obtaining stable solutions. 
A key advantage of the proposed framework is its flexibility: the model can swap priors, update rules, and projection constraints without changing the overall variational formulation.

Although the present evaluation was conducted on a controlled synthetic dataset generated according to the proposed generative model, future work will investigate the performance of the framework on real multisubject fMRI datasets and under model-mismatch scenarios, where additional sources of variability such as motion artifacts, scanner effects, and missing observations are present. Additionally, we will explore more robust and model-agnostic initialization strategies for the inference network to reduce the dependence on the GICA results, together with better-balanced optimization schemes that give the latent regularizers a stronger role during training.

\bibliographystyle{IEEEtran}
\bibliography{ref,refs_ricardo}

\end{document}